\documentclass{style}

\usepackage{lmodern}

\usepackage{amsmath,amssymb,mathtools}

\usepackage{xurl}
\usepackage{soul}
\usepackage{tabularx,wrapfig}
\usepackage{float}
\usepackage{needspace}

\hypersetup{
  pdftitle={PhySGA: A Physics-Space Grounded Agentic Framework for Generalizable and Interactive 3D Composition},
  pdfauthor={Weichen Lyu, Chong Xia, Mingyu Dong, Long Xu, Shengjun Zhang, Yueqi Duan}
}

\title{PhySGA: A Physics-Space Grounded\\
Agentic Framework for Generalizable and\\
Interactive 3D Composition}
\renewcommand{\authorfont}{\fontsize{12}{17}\selectfont}
\author{Weichen Lyu$^{1,*}$, Chong Xia$^{1,*}$, Mingyu Dong$^{1}$,\\
Long Xu$^{2}$, Shengjun Zhang$^{1}$, Yueqi Duan$^{1,\dagger}$}
\affiliation[1]{Tsinghua University}
\affiliation[2]{Zhejiang University}

\abstract{
3D object composition is pivotal to digital content creation and robot simulation, while also serving as a critical testbed for physical intelligence in foundation models. Previous methods either leverage large language models to directly predict object coordinates, or utilize generative diffusion priors to synthesize composite images for visual alignment, both of which rely on static priors in digital spaces. However, these methods consistently struggle with complex compositions such as precise hanging, gravity-driven settling and hidden containment, as these tasks require interactive construction processes involving global planning, fine-grained alignment, and physical simulation. To address these challenges, we propose \textbf{PhySGA}, a \textbf{Phy}sics-\textbf{S}pace \textbf{G}rounded \textbf{A}gentic Framework for 3D composition via agent exploration and physical simulation, enabling the universal interactive assembly of rigid, articulated, and deformable objects. Directly applying vanilla LLMs or VLMs to this task proves insufficient, as current models inherently lack grounded physical understanding and interactive construction experience. To overcome this limitation, we design a collaborative feedforward-feedback architecture: an Assembly Agent focuses on geometric and kinematic reasoning for global step planning and coarse adjustment, while a Critic Agent reasons over contact dynamics for simulator state diagnosis and fine-grained closed-loop correction. Furthermore, Action Routers map semantic planning and analysis onto executable physical primitives within the simulator, while a task-specific Self-Evolving Skill Library distills historical trial-and-error feedback into reusable experiences, boosting success rates and minimizing exploration attempts. Extensive experiments within Isaac Sim demonstrate that PhySGA significantly surpasses existing methods in relation fidelity, physical plausibility, and visual quality.
}

\setpreabstractcontent{%
  \begin{minipage}{\linewidth}
    \centering
    \includegraphics[width=1\linewidth]{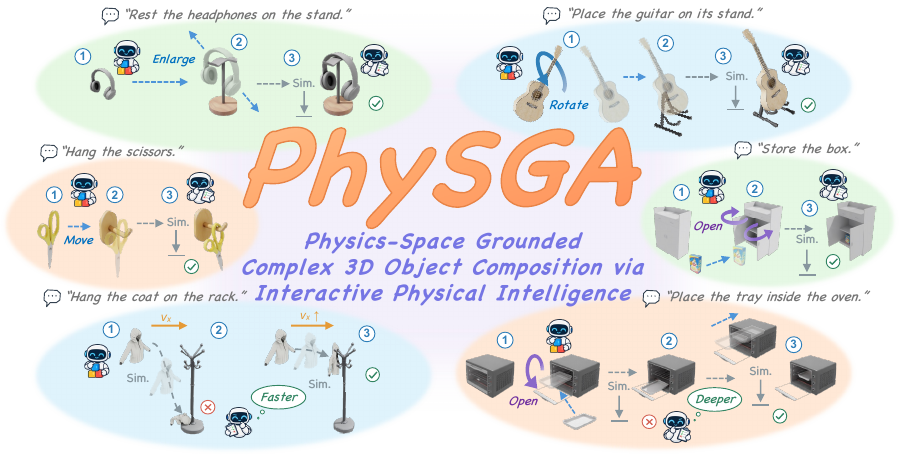}
    \captionsetup{font=footnotesize,skip=5pt}
    \captionof{figure}{\textbf{Physics-Space Grounded Complex 3D Object Composition via Interactive Physical
Intelligence. } Our framework tackles complex 3D composition by grounding the construction process
directly in physical space. Driven by a collaborative Feedforward-Feedback paradigm, the agents
iteratively propose kinematic actions, observe simulated outcomes, and apply fine-grained corrections.
This continuous loop guarantees strict physical validity and realizes complex target relations across
rigid, articulated, and deformable objects.}
    \label{fig:teaser}
  \end{minipage}%
}

\checkdata[
\raisebox{-0.2em}{\includegraphics[width=0.025\linewidth]{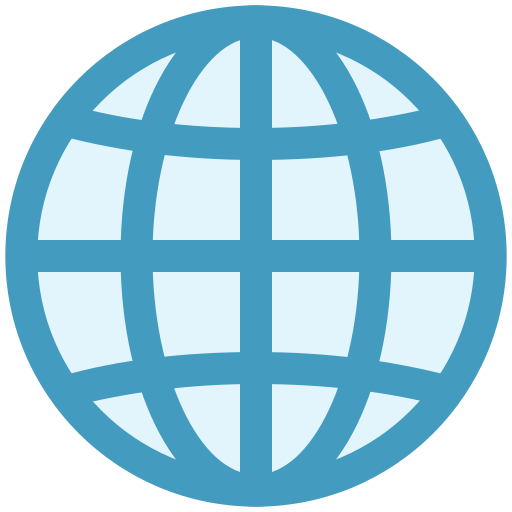}}~~Project Page]{\href{https://agenticei.github.io/PhySGA-page/}{\texttt{https://agenticei.github.io/PhySGA-page/}}
\\[-1.5ex]}

\begin{document}
\begingroup
\renewcommand{\thefootnote}{}
\footnotetext{$*$ Equal contribution, $\dagger$ Corresponding author}
\endgroup
\maketitle

\section{Introduction}
\label{sec:intro}

3D object composition aims to construct physically and functionally coherent interactions among independent object assets~\citep{qi2025twobytwo}. Empowering AI agents to construct complex interactive tasks within dynamic simulators is of paramount importance, as it yields scalable, physically grounded environments that directly enhance the capabilities of embodied models~\citep{yang2023holodeck}. Furthermore, this interactive construction establishes a rigorous standard for evaluating the true physical intelligence of foundational models. Unlike conventional benchmarks that rely on static visual question answering~\citep{fu2024blink,chow2025physbench}, 3D composition forces models to translate semantic understanding into executable physical actions, using direct simulation feedback to explicitly validate their reasoning capabilities.

Earlier scene synthesis methods place objects using predefined rules or heuristic spatial relations extracted from annotated 3D datasets~\citep{deitke2022procthor}. Recently, foundation models have driven two dominant routes for 3D composition. The language-driven route uses LLMs to directly predict object coordinates from text instructions~\citep{feng2023layoutgpt}. Agentic frameworks like SAGE~\citep{xia2026sage} and SceneSmith~\citep{pfaff2026scenesmith} extend this approach by employing tool-use to iteratively adjust hierarchical placements. Yet, their core mechanisms remain heavily dependent on semantic layout priors. The vision-driven route relies on 2D generative models to create an image of the target composition. This image subsequently guides the spatial positioning of 3D objects using diffusion priors~\citep{shan2026interact3d}.

However, these methods primarily rely on static priors in digital space, namely knowledge of object positions and relations encoded in language models or diffusion priors, struggling with complex 3D object compositions (e.g., precise hanging, gravity-driven settling, and hidden containment). This is because precise hanging requires precise multi-view observation and exact spatial alignment; gravity-driven settling must explicitly utilize physical simulation under gravitational dynamics; and hidden containment involves a sequence of interactive kinematic actions. Looking back at these approaches, the language-driven branch struggles in these scenarios because it learns plausible static layouts but lacks experiential data for complex interactions, while the final state of physical equilibrium is notoriously difficult to express precisely in text. Equally problematic, the vision-driven branch prioritizes visual plausibility, which rarely equates to physical validity, proving fatal for fine-grained assembly. Therefore, a fundamental challenge in complex interactions is realizing target relations while strictly adhering to physical laws. We argue that a reliable approach is to ground the composition directly in physical space, assembling objects step-by-step in an iterative manner. This preserves physical validity at each step and ultimately ensures relation satisfaction. Intuitively, this mirrors how humans assemble objects in the real world: not through a one-shot prediction, but through a continuous loop of planning, adjusting, and utilizing physical feedback.

In this paper, we propose PhySGA, formulating 3D object composition as an interactive assembly process in physical space, inspired by how humans assemble objects in the real world, as illustrated in Figure~\ref{fig:teaser}. To achieve this, we design a collaborative Feedforward-Feedback architecture tailored for physical interaction. The Assembly Agent conducts kinematic reasoning to formulate a global plan, executing metric-grounded coarse adjustments to achieve a reliable preliminary correction. Subsequently, the Critic Agent analyzes simulated motion and contact feedback to diagnose physical failures, performing pixel-grounded fine-grained corrections until the target relation reaches physical equilibrium. To bridge high-level semantic reasoning and low-level simulation execution, we introduce modular Action Routers that seamlessly convert agent predictions into exact geometric, articulated, and velocity commands. Finally, we maintain a Self-Evolving Skill Library that distills historical interaction memories, allowing the agents to retrieve past experience to avoid repeating failures and facilitate continuous experience reuse.

Our main contributions are summarized as follows:
\begin{itemize}
  \item We propose PhySGA, formulating 3D object composition as an interactive assembly process grounded in dynamic physical space, ensuring that complex target relations are constructed upon strict physical validity.
  \item We design a comprehensive agentic framework featuring a Feedforward-Feedback paradigm driven by an Assembly Agent and a Critic Agent for global planning and precise visual correction, modular Action Routers for executing decoupled physical commands, and a Self-Evolving Skill Library for historical experience reuse.
  \item Extensive experiments in Isaac Sim across rigid, articulated, and deformable tasks demonstrate that PhySGA significantly outperforms state-of-the-art methods, providing a scalable solution for complex task construction in simulators while effectively unlocking the latent physical intelligence of foundational models.
\end{itemize}

\section{Related Work}
\label{sec:related}

\subsection{Simulation Scene Generation}

Simulation scene generation provides interactive environments for training embodied agents. Early approaches construct these environments through procedural rules~\citep{deitke2022procthor,raistrick2024infinigenindoors,raistrick2023infinigen} or by learning distributions over static object layouts~\citep{paschalidou2021atiss,tang2023diffuscene,wang2019planit,ritchie2019fastsynth,wang2021sceneformer}. Recent language-guided methods, such as Holodeck and SceneWeaver~\citep{yang2023holodeck,yang2025sceneweaver}, advance this field by directly translating textual instructions into object selections and spatial arrangements~\citep{lin2024instructscene}. Language agents can interleave reasoning with actions and use environmental observations to guide subsequent decisions~\citep{yao2023react}. To further refine these layouts, agentic frameworks like SAGE~\citep{xia2026sage} and SceneSmith~\citep{pfaff2026scenesmith} incorporate iterative tool use alongside visual or physical critics. However, these language-driven pipelines primarily formulate generation as predicting static coordinates, treating physical simulation merely as a binary verification step. They lack the capacity to handle complex relations that require continuous interaction. To address this, PhySGA bypasses one-shot coordinate prediction by embedding the agent within the simulator, driving composition through step-by-step physical execution.

\subsection{3D Object Composition}

3D object composition aims to establish functional and spatial relations among distinct 3D objects. Previous research explores spatial arrangements across different scales. At the \textit{scene} level, methods like SimRecon~\citep{xia2026simrecon} and PhyRecon~\citep{ni2024phyrecon} reconstruct 3D scenes from visual observations~\citep{dong2026replicateanyscene,lee2026simuscene,li2023rico,wu2023objectsdf,ni2025dprecon,xia2026onlinex}. At the \textit{part} level, methods such as Jigsaw~\citep{lu2023jigsaw} assemble fractured pieces~\citep{sellan2022breakingbad} by matching local geometries. Shifting the focus toward complete \textit{object composition}, Interact3D~\citep{shan2026interact3d} leverages diffusion priors to guide the geometric registration of independent assets. However, these approaches rely heavily on visual alignment to recover static terminal poses, which does not guarantee physical plausibility. Crucially, pure visual alignment completely fails on complex interactions: gravity-driven settling dictates dynamic physical simulation to determine the final resting state, and hidden containment involves visually occluded target regions, rendering image-based matching ineffective. PhySGA overcomes these limitations by casting composition as a closed-loop physical construction problem, ensuring that object relations are rigorously validated by physical laws.

\subsection{Evaluating Physical Intelligence}

Evaluating physical intelligence in foundation models has increasingly drawn attention~\citep{chow2025physbench}. Current benchmarks primarily probe this capability through visual question answering, requiring models to infer spatial relations~\citep{fu2024blink,ramakrishnan2025space}, metric quantities~\citep{yang2025thinking}, or physical dynamics~\citep{yi2020clevrer,chen2022comphy} from static images or videos~\citep{song2025siribench,chow2025physbench}. Concurrently, experiential language agents demonstrate that models can improve over time by accumulating trial-and-error feedback~\citep{majumder2024clin,chen2024automanual,shinn2023reflexion,zhao2023expel} or self-evolving skills across episodes~\citep{wang2023voyager,xia2026skillrl,zhao2024learnact}. However, passive visual questioning is often insufficient to capture the complexities of real-world physics. The most rigorous testbed for physical intelligence is executable construction within a simulator, where semantic reasoning must be validated by dynamic physical interaction. PhySGA adopts this paradigm, evaluating and enhancing the physical understanding of agents through continuous simulation feedback rather than static predictions.

\section{Method}
\label{sec:method}

\begin{figure}[t]
  \centering
  \includegraphics[width=\linewidth]{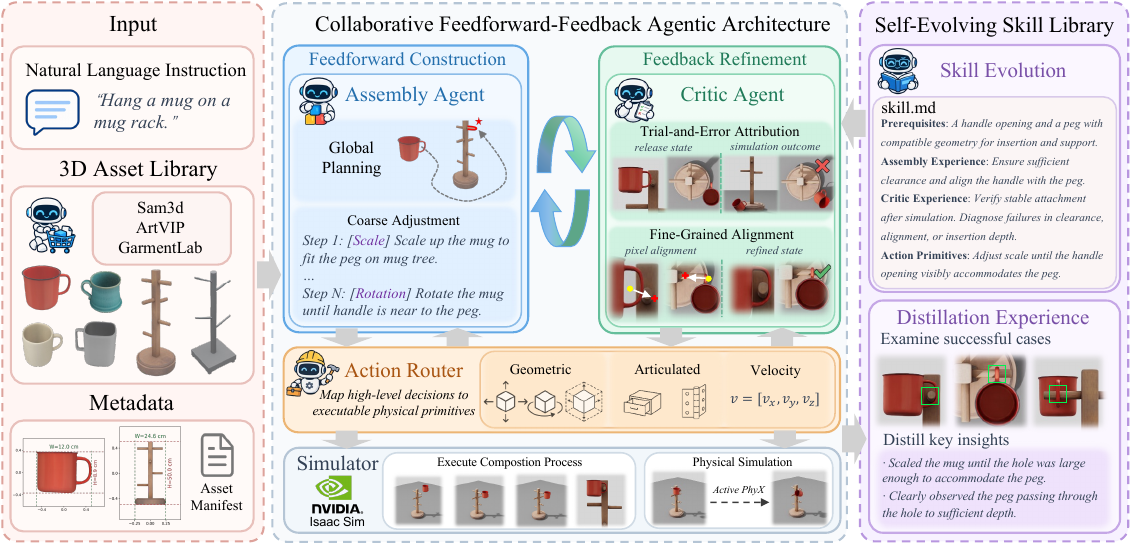}
  \caption{\textbf{Overview of the PhySGA framework.} Driven by a collaborative Feedforward-Feedback architecture, the Assembly Agent executes metric-grounded coarse adjustments, while the Critic Agent performs pixel-grounded fine-grained corrections via simulation feedback. Action Routers translate these semantic decisions into decoupled physical commands, and a Self-Evolving Skill Library distills successful interactions into reusable skills.}
  \label{fig:physga_pipeline}
\end{figure}

We present PhySGA, a physics-space grounded agentic framework for interactive 3D composition, as illustrated in Figure~\ref{fig:physga_pipeline}. Section~\ref{sec:method_overview} presents the task formulation, the overall architecture, and the unified asset pipeline. Section~\ref{sec:high_level_agents} introduces the collaborative high-level agents, while Sections~\ref{sec:action_router} and \ref{sec:self_evolving_skills} detail the modular action routing and self-evolving skills, respectively.

\subsection{Overview of PhySGA}
\label{sec:method_overview}

\subsubsection{Interactive Composition Representation}
\label{sec:composition_representation}

Given a natural-language instruction $q$ and a pair of independently instantiated, simulation-ready 3D assets $\mathcal{A}=\{a_1,a_2\}$, the goal is to construct a physically realizable composition $\mathcal{C}^{*}$ from an initial simulator state $s_0^*$. Previous methods predominantly formulate this task as a one-shot prediction of final relative object poses, bypassing the physical interaction trajectory. However, complex compositions inherently depend on continuous contact dynamics and intermediate state transitions. We therefore formulate 3D object composition as an iterative physical construction process. Starting from $s_0^*$, the composition is achieved through a sequence of state transitions utilizing a comprehensive set of operations, including geometric transformations (translation, rotation, scale), dynamic velocity assignments, and physical simulation. This generalized iterative representation can be written as:
\begin{equation}
  s_{k} = \Phi(s_{k-1}, a_k),
  \label{eq:general_iterative}
\end{equation}
where $a_k$ represents an applied adjustment or simulation step at iteration $k$, and $\Phi$ denotes the environment transition function. This step-by-step evolution continues until it yields the final target composition $\mathcal{C}^* = s_K$.

\subsubsection{Feedforward-Feedback Agentic Architecture}
\label{sec:agentic_architecture}

To execute this generalized iterative process, we design a hierarchical architecture comprising two high-level agents. The Assembly Agent ($\pi_{\mathrm{Asm}}$) governs the outer loop of feedforward construction, leveraging geometric and kinematic reasoning to formulate $N$-step plans and execute coarse metric adjustments. For the $n$-th step, it applies adjustments to the previously validated state $s_{n-1}^*$, producing a new state $s_n$. If this step requires physical validation, $s_n$ initiates a pre-simulation state $s_{n,0}$. The Critic Agent ($\pi_{\mathrm{Crit}}$) manages the conditional inner loop of feedback refinement. During the $r$-th round, the simulator $\mathcal{S}$ evaluates this state, and the Critic Agent applies fine-grained corrections based on the simulated physical evidence to produce the revised state $s_{n,r+1}$. Once the physical requirements are satisfied, the validated state returns to the feedforward process as $s_n^*$. The final target composition is achieved upon the completion of all $N$ steps, yielding $\mathcal{C}^{*} = s_N^*$. The state evolution for step $n$ is explicitly instantiated as:
\begin{equation}
  \begin{aligned}
    &s_n = \pi_{\mathrm{Asm}}(q,\mathcal{A},s_{n-1}^*), \\
    &s_n \to s_{n,0} \quad \text{(if physical validation is required)}, \\
    &s_{n,r+1} = \pi_{\mathrm{Crit}}\!\left(q,\mathcal{A},s_{n,r},\mathcal{S}(s_{n,r})\right), \\
    &s_{n,r+1} \to s_n^* \quad \text{(upon validation success)}.
  \end{aligned}
  \label{eq:agentic_architecture}
\end{equation}
This asymmetric collaboration structurally separates the step-wise global assembly planning from the pixel-grounded physical correction within the simulator.

\subsubsection{Unified Asset Pipeline}
\label{sec:asset_construction}

PhySGA processes generated rigid assets using GPT Image 2~\citep{openai2026gptimage2} and SAM 3D Objects~\citep{sam3dteam2025sam3d} alongside articulated and deformable assets retrieved from ArtVIP~\citep{jin2026artvip} and GarmentLab~\citep{lu2024garmentlab} through a unified pipeline. To provide the high-level agents with consistent observations across diverse asset types, we standardize all assets using a shared metadata schema that explicitly records semantics, metric geometry, canonical orientation, and physical properties. We additionally provide standardized CAD-style orthographic multi-view renderings with metric annotations to facilitate accurate spatial reasoning.

\subsection{Collaborative Dual-Agent Architecture}
\label{sec:high_level_agents}

\subsubsection{Shared State Machine Design}
\label{sec:shared_agent_state_machine}

To effectively explore the interactive assembly process, we structure both the Assembly and Critic agents as persistent state machines. Unlike standard one-shot coordinate prediction, this cyclic formulation endows the agents with temporal awareness and environment-conditioned reasoning capabilities. The interaction cycle unfolds through a unified sequence of four cognitive phases: \textsc{Observe}, \textsc{Decide}, \textsc{Execute}, and \textsc{Validate}. Formally, at any interaction timestep $t$, an active agent $i \in \{\mathrm{Asm}, \mathrm{Crit}\}$ begins in the \textsc{Observe} phase, processing the current simulator observation $o_t$ alongside its historical memory $\mathcal{M}_{t-1}^i$. Based on this context, the agent transitions to the \textsc{Decide} phase to formulate a geometric or physical adjustment, outputting a natural-language instruction $d_t^i$. During the \textsc{Execute} phase, the Action Router grounds this semantic instruction into a concrete simulator command, yielding a physical execution status. Finally, in the \textsc{Validate} phase, the agent evaluates the outcome. At this juncture, the agent determines the macroscopic control flow: advancing to the next step if the current requirements are satisfied, or looping back to initiate another attempt if they are not. Regardless of success or failure, the executed action and its environmental consequence are immediately assimilated into the current memory $\mathcal{M}_t^i$ to inform the subsequent reasoning cycle.

\subsubsection{Assembly Agent}
\label{sec:assembly_agent}
\paragraph{Global Planning.} To emulate the human cognitive process of feedforward task decomposition, we equip the Assembly Agent with a structured chain-of-thought prior~\citep{wei2022chainofthought}. Starting from the initial simulator state $s_0$ and the intended assembly relation given by the task instruction, it generates a natural language plan that outlines the whole construction process. Each step provides a semantic description of a target object state and its observable check condition. Given the task instruction $q$, task-relevant asset semantics $\eta_{\mathcal{A}}$, the initial observation $o_0$, and its memory $\mathcal{M}_{\mathrm{Asm}}$, it produces:
\begin{equation}
  \mathcal{P}_{\mathrm{Asm}}
  = \operatorname{Plan}_{\mathrm{Asm}}
  \left(q,\eta_{\mathcal{A}},o_0,\mathcal{M}_{\mathrm{Asm}}\right)
  = \left\{(g_k,c_k)\right\}_{k=1}^{K},
  \qquad K\leq K_{\max}.
  \label{eq:assembly_agent}
\end{equation}
Each pair $(g_k,c_k)$ serves as a high-level semantic instruction that specifies the target state and verification criteria, establishing a global trajectory rather than prescribing a rigid sequence of low-level commands.

\paragraph{Metric-Grounded Coarse Adjustment.} To execute these planned steps, the Assembly Agent transmits the intended adjustment type, the estimated metric value, the target state, and the corresponding verification criteria to its designated Action Router. Operating within an observation space of orthogonal multi-view scene renderings, the agent selects one or more primary views that best expose the interactive region. It then reasons over these selected views to deduce the actual metric values required for the adjustment, thereby further concretizing the target state and its checking criteria. This metric-output paradigm maximally preserves the inherent common-sense spatial reasoning capabilities of vision-language models, making it highly effective for coarse global coordination and structural alignment.

\subsubsection{Critic Agent}
\label{sec:critic_agent}
\paragraph{Trial-and-Error Attribution.} In contrast to the Assembly Agent's feedforward approach, the Critic Agent emulates human trial-and-error refinement based on physical feedback. Following the $r$-th physical simulation round of step $n$, the Critic Agent diagnoses the failure $\xi_{n,r}$ and identifies a correction aspect $\alpha_{n,r}$ from the simulated state $s_{n,r}^{\mathrm{sim}}$. Using this diagnosis alongside the latest observations, it refines the pre-simulation state $s_{n,r}$ to produce an adjusted state $s_{n,r+1}$, which is then evaluated by a subsequent physical simulation. If the resulting simulation succeeds, $s_{n,r+1}$ is returned as the validated state. If it fails, the agent strategically decides whether to proceed with further refinements from the newly adjusted state $s_{n,r+1}$ or to roll back to the previous state $s_{n,r}$ for an alternative correction path. This physical refinement process is formulated as:
\begin{equation}
  \begin{aligned}
    &(\xi_{n,r},\alpha_{n,r}) = \operatorname{Diagnose}_{\mathrm{Crit}}
    \left(q, s_{n,r}^{\mathrm{sim}}, \mathcal{M}_{\mathrm{Crit},r}\right), \\
    &s_{n,r+1} = \operatorname{Refine}_{\mathrm{Crit}}
    \left(s_{n,r}, o_{n,r}, \xi_{n,r}, \alpha_{n,r}, \mathcal{M}_{\mathrm{Crit},r}\right).
  \end{aligned}
  \label{eq:critic_agent}
\end{equation}

\paragraph{Pixel-Grounded Fine-Grained Alignment.} Coarse metric adjustments are often insufficient for achieving precise physical alignment. To address this, we directly leverage the inherent pixel-level alignment capabilities of Vision-Language Models. During the diagnosis phase following a failed simulation, the Critic Agent examines the multi-view observations to identify the specific geometric error. Rather than estimating semantic metric values, the agent directly outputs pixel coordinates, mapping the representative pixel of the manipulated object to its intended position on the target object across multiple views. The Action Router then triangulates these pixel correspondences using calibrated camera projections, converting the precise visual alignment directly into executable commands for the simulator.

\subsection{Action Router}
\label{sec:action_router}
The Action Router grounds the high-level agents' semantic targets and adjustment analyses into executable simulator commands through three modular functions. The \textbf{Geometric Router} converts translation, rotation, and uniform scale predictions along 3D orthogonal axes into exact transformations. The \textbf{Articulated Router} maps semantic instructions (e.g., opening a drawer) to corresponding joint movements for articulated objects. The \textbf{Velocity Router} converts release instructions (e.g., throwing forward) into uniform world-space velocities applied to deformable objects prior to simulation. This design seamlessly adapts to diverse object types and explicitly decouples distinct physical variables. Moreover, by separating high-level semantic reasoning from low-level physical details, it significantly reduces the cognitive burden on the Assembly Agent and Critic Agent, allowing them to focus exclusively on task planning and spatial alignment.

\subsection{Self-Evolving Skill Library}
\label{sec:self_evolving_skills}
To avoid repeating past failures, reduce reasoning time, and streamline planning for tasks with similar physical conditions, we maintain a self-evolving skill library $\mathcal{K}$. Before executing a new instruction $q$, the framework retrieves compatible knowledge from $\mathcal{K}$ to provide the agents with reference strategies. After execution, the consolidated interaction memory $\mathcal{M}$ and final physical outcome $\mathcal{O}$ are jointly distilled to construct new knowledge or supplement existing skills. This distillation summarizes key insights, such as Assembly planning bottlenecks and the Critic's root-cause error analyses. The entire retrieval and evolution process is concisely formulated as:
\begin{equation}
  \begin{aligned}
    p_i &\leftarrow p_i \oplus \operatorname{Retrieve}(\mathcal{K}, q), \\
    \mathcal{K} &\leftarrow \operatorname{Update}\big(\mathcal{K}, \operatorname{Distill}(\mathcal{M}, \mathcal{O})\big),
  \end{aligned}
  \label{eq:self_evolving_skills}
\end{equation}
where $\oplus$ denotes appending the retrieved skill to the prompt $p_i$ of agent $i \in \{\mathrm{Asm}, \mathrm{Crit}\}$.

\section{Experiments}
\label{sec:experiments}

We conduct extensive experiments to evaluate the effectiveness of our method in 3D composition. Following the experimental setup in Section~\ref{sec:experimental_setup}, we compare PhySGA against state-of-the-art 3D composition frameworks in Section~\ref{sec:baseline_comparison} and vanilla coding agents across various base models in Section~\ref{sec:base_model_comparison}. Finally, Section~\ref{sec:agent_ablations} provides comprehensive ablations of our core agentic components.

\subsection{Experimental Setup}
\label{sec:experimental_setup}

\paragraph{Benchmark.}
We construct a benchmark of object composition tasks spanning three categories: rigid, articulated, and deformable. Rigid tasks cover physical contact relations (e.g., containment, slotting) between rigid objects. Articulated tasks involve the actuation of kinematic joints, while deformable tasks require physical simulation to naturally settle soft bodies. The benchmark comprises 40 unique tasks (20 rigid, 10 articulated, 10 deformable). Each task is instantiated with three prompt variants featuring different asset descriptions but identical target relations, yielding 120 evaluation scenes in total.

\paragraph{Baselines.}
We compare PhySGA against three representative systems: SAGE~\citep{xia2026sage}, SceneSmith~\citep{pfaff2026scenesmith}, and Interact3D~\citep{shan2026interact3d}. Since SAGE and SceneSmith are hierarchical scene generation frameworks, we primarily compare against their object-level performance. Specifically, we retain the placement and critic agents from SAGE and the manipuland agents from SceneSmith, while omitting room initialization and global layout planning. Each system utilizes its native asset acquisition pipeline, and all final compositions are executed and evaluated in Isaac Sim under a unified protocol.

\paragraph{Implementation Details.}
All methods are executed and evaluated in NVIDIA Isaac Sim 4.2.0. In the baseline comparison, GPT-5.5 serves as the base VLM for all VLM-driven reasoning. Each method receives the same 120 language prompts and follows its native pipeline to generate compositions, which are then loaded into Isaac Sim using identical rendering and physical execution configurations.

\paragraph{Metrics.}
We evaluate composition quality using both physical and visual metrics. Physical metrics include Penetration Rate (PR) and Contact Rate (CR), both ranging from 0 to 1, which measure the average fraction of scenes exhibiting invalid collisions (lower is better) and successful target contacts (higher is better), respectively. For visual metrics, a VLM evaluator scores the multi-view outputs on a scale from 0 to 10. We report Asset Fidelity Score (AFS) for asset fidelity, Composition Overall Score (COS) for overall composition quality, and prominently highlight Relation Fidelity Score (RFS) as the primary indicator of successfully achieving the instructed relation. All visual metrics are averaged across the 120 scenes.

\subsection{Comparison with Baselines}
\label{sec:baseline_comparison}

\begin{figure*}[t]
  \centering
  \includegraphics[width=\linewidth]{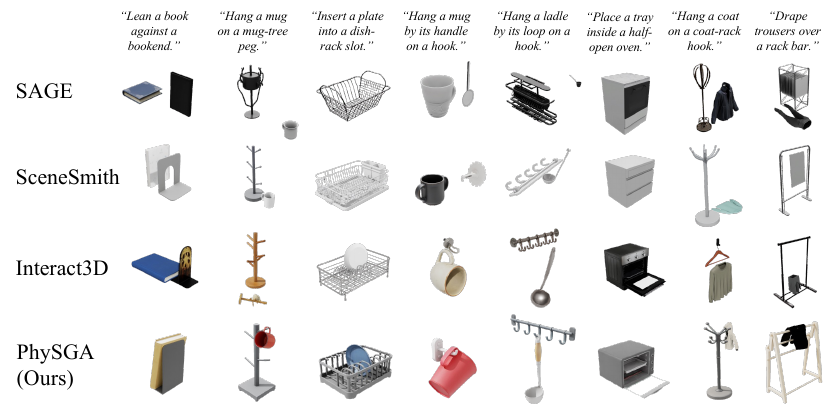}
  \caption{\textbf{Qualitative comparison with baselines.} Columns represent distinct compositional tasks across rigid, articulated, and deformable domains, comparing SAGE, SceneSmith, Interact3D, and our PhySGA after physical simulation. PhySGA consistently maintains high-fidelity asset details, satisfies target relations, and yields physically valid compositions.}
  \label{fig:baseline_qualitative}
\end{figure*}

\begingroup
\begin{wraptable}{r}{0.60\linewidth}
  \vspace{-\intextsep}
  \setlength{\abovecaptionskip}{0pt}
  \centering
  \caption{\textbf{Quantitative comparison with existing methods.}
  PR and CR range from 0 to 1, while AFS, RFS, and COS range from 0 to 10.
  The best and second-best results are shown in bold and underlined, respectively.}
  \label{tab:baseline_comparison}
  \vspace{4pt}

  \footnotesize
  \setlength{\tabcolsep}{2pt}
  \renewcommand{\arraystretch}{1.06}
  \renewcommand{\tabularxcolumn}[1]{m{#1}}

\begin{tabularx}{\linewidth}{
  >{\raggedright\arraybackslash}m{0.36\linewidth}|
  *{2}{>{\centering\arraybackslash}X}|
  *{3}{>{\centering\arraybackslash}X}
}
    \toprule
    \multirow{2}{*}{Method}
      & \multicolumn{2}{c|}{Physical Metrics}
      & \multicolumn{3}{c}{Visual Metrics} \\
    & PR$\downarrow$ & CR$\uparrow$
    & AFS$\uparrow$ & RFS$\uparrow$ & COS$\uparrow$ \\
    \midrule
    SAGE~\citep{xia2026sage}
      & \textbf{0.000} & 0.000 & 3.96 & 0.74 & 3.03 \\
    SceneSmith~\citep{pfaff2026scenesmith}
      & 0.042 & 0.575 & 3.29 & 2.59 & 4.55 \\
    Interact3D~\citep{shan2026interact3d}
      & 0.133 & \underline{0.717} & \underline{6.87}
      & \underline{4.45} & \underline{5.12} \\
    \midrule
    PhySGA (Ours)
      & \underline{0.033} & \textbf{0.958} & \textbf{9.02}
      & \textbf{8.75} & \textbf{7.41} \\
    \bottomrule
  \end{tabularx}
\end{wraptable}

Table~\ref{tab:baseline_comparison} presents the quantitative evaluation
of the composition tasks. PhySGA significantly outperforms all baselines
across both physical and visual metrics, achieving a near-perfect Contact
Rate (CR) of 0.958 while maintaining a critically low Penetration Rate
(PR) of 0.033. Notably, SAGE records a PR of 0.000, but this is merely an
artifact of its inability to establish interaction. It places assets far
apart without meaningful contact, resulting in a 0.000 CR and the lowest
Relation Fidelity Score (RFS). In contrast, PhySGA's RFS of 8.75 nearly
doubles that of the strongest baseline, Interact3D at 4.45, demonstrating
its superior capability to translate semantic relations into physically
validated configurations.

Qualitative comparisons in Figure~\ref{fig:baseline_qualitative} further
highlight these advantages. Baseline methods frequently struggle with
spatial tolerances, resulting in floating assets, as observed with SAGE,
or severe visual misalignments after physical simulation, as observed
with Interact3D. Conversely, PhySGA consistently produces physically
coherent compositions across all object categories. It uniquely succeeds
in fine-grained operations requiring millimeter-level precision, such as
inserting a plate into a narrow dish-rack slot or hanging a ladle by its
loop. Furthermore, it effectively manages complex multi-step articulated
interactions and realistic deformable settling in simulation.

\WFclear
\endgroup
\enlargethispage{\baselineskip}

\Needspace{0.48\textheight}
\subsection{Comparison with Vanilla VLM-based Coding Agents}
\label{sec:base_model_comparison}

We compare vanilla VLM-based coding agents against the same base models instantiated within PhySGA, evaluating Qwen3.8-Max~\citep{alibaba2026qwen38max}, Gemini-3.1-Pro-Preview~\citep{google2026gemini31pro}, DeepSeek-V4-Flash-Vision-Exp~\citep{deepseek2026v4flashvision}, and GPT-5.5~\citep{openai2026gpt55}. All configurations share identical benchmark tasks and asset instances. Therefore, we omit the Asset Fidelity Score (AFS) and focus exclusively on physical and relational metrics.

\begingroup

\begin{wraptable}{r}{0.60\linewidth}
  \vspace{-\intextsep}
  \setlength{\abovecaptionskip}{0pt}
  \centering
  \caption{\textbf{Comparison between vanilla VLM-based coding agents and PhySGA using the same base VLMs.} The best and second-best results are shown in bold and underlined, respectively.}
  \label{tab:base_model_comparison}
  \vspace{4pt}
  \footnotesize
  \setlength{\tabcolsep}{2.5pt}
  \renewcommand{\arraystretch}{1.06}
  \renewcommand{\tabularxcolumn}[1]{m{#1}}
  \begin{tabularx}{\linewidth}{>{\raggedright\arraybackslash}X|*{2}{>{\centering\arraybackslash}m{0.105\linewidth}}|*{2}{>{\centering\arraybackslash}m{0.105\linewidth}}}
    \toprule
    \multirow{2}{*}{Method} & \multicolumn{2}{c|}{Physical Metrics} & \multicolumn{2}{c}{Visual Metrics} \\
    & PR$\downarrow$ & CR$\uparrow$ & RFS$\uparrow$ & COS$\uparrow$ \\
    \midrule
    Qwen3.8-Max & 0.258 & 0.542 & 1.42 & 2.46 \\
    DeepSeek-V4 & 0.208 & 0.542 & 1.16 & 2.35 \\
    Gemini-3.1-Pro & 0.158 & 0.392 & 1.79 & 3.08 \\
    GPT-5.5 & 0.142 & 0.350 & 1.29 & 2.62 \\
    \midrule
    PhySGA + Qwen3.8-Max & 0.092 & \underline{0.858} & \underline{6.36} & \underline{6.39} \\
    PhySGA + DeepSeek-V4 & 0.075 & 0.467 & 4.11 & 4.72 \\
    PhySGA + Gemini-3.1-Pro & \underline{0.058} & 0.792 & 5.90 & 6.06 \\
    PhySGA + GPT-5.5 & \textbf{0.033} & \textbf{0.958} & \textbf{8.75} & \textbf{7.41} \\
    \bottomrule
  \end{tabularx}
\end{wraptable}

The vanilla baselines operate as Codex-style coding agents, utilizing a Python sandbox to analyze asset geometry and generate executable scene programs. Table~\ref{tab:base_model_comparison} shows that these vanilla agents struggle to realize physically plausible configurations, highlighting an inherent deficit in their native physical intelligence. Conversely, integrating these models into PhySGA consistently reduces penetration and drastically improves relation fidelity across all models. GPT-5.5 achieves the highest overall performance, followed by Qwen3.8-Max. This confirms that our framework effectively unlocks and empowers the latent physical intelligence of foundational models for complex 3D composition.

\WFclear
\endgroup
\Needspace{0.42\textheight}
\subsection{Ablation Study}
\label{sec:agent_ablations}

\begingroup
\begin{wraptable}{r}{0.60\linewidth}
  \vspace{-\intextsep}
  \setlength{\abovecaptionskip}{0pt}
  \centering
  \caption{\textbf{Ablation study of PhySGA with individual components removed.} PhySGA denotes the complete framework with all components enabled.}
  \label{tab:agent_ablations}
  \vspace{4pt}
  \footnotesize
  \setlength{\tabcolsep}{2.5pt}
  \renewcommand{\arraystretch}{1.06}
  \renewcommand{\tabularxcolumn}[1]{m{#1}}
  \begin{tabularx}{\linewidth}{>{\raggedright\arraybackslash}X|*{2}{>{\centering\arraybackslash}m{0.105\linewidth}}|*{2}{>{\centering\arraybackslash}m{0.105\linewidth}}}
    \toprule
    \multirow{2}{*}{Method} & \multicolumn{2}{c|}{Physical Metrics} & \multicolumn{2}{c}{Visual Metrics} \\
    & PR$\downarrow$ & CR$\uparrow$ & RFS$\uparrow$ & COS$\uparrow$ \\
    \midrule
    \multicolumn{1}{l|}{\textit{Assembly Agent}} & \multicolumn{2}{c|}{} & \multicolumn{2}{c}{} \\
    w/o Global Planning &0.067 &0.733 &6.08 &6.42 \\
    w/o Metric Adjustment &0.050 &0.783 &6.92 &6.58 \\
    \midrule
    \multicolumn{1}{l|}{\textit{Critic Agent}} & \multicolumn{2}{c|}{} & \multicolumn{2}{c}{} \\
    w/o Error Attribution &0.092 &\underline{0.917} &7.42 &7.08 \\
    w/o Pixel Alignment &0.083 &0.667 &6.33 &5.92 \\
    \midrule
    w/o Action Routers &0.058 &0.283 &4.12 &5.42 \\
    w/o Self-Evolving Skills &\underline{0.042} &0.792 &\underline{7.58} &\underline{7.17} \\
    \midrule
    PhySGA &\textbf{0.033} &\textbf{0.958} &\textbf{8.75} &\textbf{7.41} \\
    \bottomrule
  \end{tabularx}
\end{wraptable}

We evaluate six ablations of PhySGA under the same experimental settings to assess the contribution of each core component to overall composition quality. \textit{w/o Global Planning} replaces the complete global plan with isolated step-by-step planning. \textit{w/o Metric Adjustment} replaces metric estimation by directly predicting pixel-level movement changes. \textit{w/o Error Attribution} performs direct adjustments without utilizing physical simulation feedback. \textit{w/o Pixel Alignment} forces the agent to directly predict real-valued spatial movements instead of precise pixel alignments. \textit{w/o Action Routers} requires the agents to output low-level simulator control codes directly without executing through modular routers. \textit{w/o Self-Evolving Skills} removes the retrieval of historical experience.

As reported in Table~\ref{tab:agent_ablations}, the complete PhySGA consistently achieves the highest Contact Rate (CR) and visual scores (RFS, COS), alongside the lowest Penetration Rate (PR). Removing any single module leads to a noticeable drop in overall composition quality. This confirms that global planning, coarse-to-fine visual alignment, modular action routing, and self-evolving skills are all indispensable for physically grounded 3D composition.

\WFclear
\endgroup

\section{Conclusion}
\label{sec:conclusion}

In this paper, we propose PhySGA, a physics-space grounded agentic framework for interactive 3D object composition across rigid, articulated, and deformable objects. We design a collaborative feedforward-feedback architecture in which the Assembly Agent uses geometric and kinematic reasoning for global assembly planning and metric-grounded coarse adjustments, while the Critic Agent diagnoses composition failures from simulated contact dynamics and performs pixel-grounded fine adjustments. Action Routers translate their decisions into executable simulator commands, and a Self-Evolving Skill Library distills experience from complete assembly runs into reusable skills for subsequent tasks. Experiments in Isaac Sim show improved relation fidelity and overall composition quality over existing methods and vanilla VLM-based coding agents, with ablation studies further supporting the effectiveness of the key design choices.

\bibliographystyle{plainnat}
\bibliography{main}

\end{document}